# A Unified Approach of Observability Analysis for Airborne SLAM

Qiang Fang

*Abstract*--An unmanned aerial vehicle(UAV) is tasked to explore an unknown environment and to map the features it finds when GPS is denied. It navigates using a statistical estimation technique known as simultaneous localization and mapping(SLAM) which allows for the simultaneous estimation of the location of the UAV as well as the location of the features it sees. Observability is a key aspect of the state estimation problem of SLAM, However, the dimension and variables of SLAM system might be changed with new features, to which little attention is paid in the previous work. In this paper, a unified approach of observability analysis for SLAM system is provided, whether the dimension and variables of SLAM system are changed or not, we can use this approach to analyze the local or total observability of the SLAM system.

*Index Terms*-- Unmanned Aerial Vehicle (UAV); Simultaneous Localization and Mapping(SLAM); Inertial Navigation System(INS); Observability; PWCS; Extend Kalman Filter(EKF)

## 1. INTRODUCTION

Future solar system exploration is filled with missions that require landing spacecraft on planets, moons, comets, and asteroids. Consider the scenario where an unmanned aerial vehicle(UAV) is tasked with building an internal representation of the location of features in an environment. In this scenario however, the UAV does not come equipped with its own self-localisation aid such as a Global Position System(GPS) receiver and does not have any a priori terrain data information from which to localize. The problem is compounded by the requirement for the UAV to have information regarding its velocity and attitude in order to stabilize its motion. In many applications the vehicle uses an Inertial Navigation System(INS), however the eventual growth of localization errors due to the dead-reckoning nature of the INS limits the time that the vehicle can operate. So the paradigm known as simultaneous localization and mapping(SLAM)[1-4] in which the vehicle pose and a feature map are estimated simultaneously using only relative observations of the locations of features with respect to the vehicle.

In practice, a filter for recursive state estimation is employed to correct the system errors using sensor aiding information. In designing the filter, the observability analysis is necessary because observability determines the existence of solutions and sets a lower limit on the estimation error.

In SLAM system, the study of the observability of such estimation processes is relevant for predicting the convergence or for taking a decision where to go in order to guarantee the localization of the platform.

Observability[5] analysis has had a long history in studying aid-INS. In [6,7] the authors present a new method of observability analysis for piece-wise constant linear systems and apply the method to the analysis of an aided-INS during in-flight alignment. More recently observability has been used to study GPS-aided INS [8,9] during sensor platform maneuvers. In [10], the authors study the number of unobservable states in the inertial SLAM algorithm by casting the equations into the indirect or error form. Based on the work above, in [11], the authors investigate which are the unobservable states and how the unobservability is related to the vehicle motions, and using information theory to select the vehicle's motion. While in [10-11], the authors do not analyze the observability of SLAM system when the dimension and variables are changed with new features which is the fact in most SLAM systems. To solve the problem, a unified approach of observability analysis for SLAM system is provided, whether the dimension and variables of SLAM system are changed or not, we can use this approach to analyze the local or total observability of the SLAM system.

This paper is organized as follows. Section II briefly discusses the error model of the inertial sensor-based SLAM system. Section III discusses the theory of continuous PWCS observability. The unified approach of SLAM observability is devoted in Section IV. In section V, four cases in SLAM are considered . In section VI, simulation results are listed. Conclusions are made in Section VII.

## 2. ERROR MODEL

The INS error equation and the SLAM observation model in continuous time can be found in [11,12]

### 2.1 INS error equation

$$\delta \dot{X}(t) = F\delta X(t) + w(t) \quad (1)$$

Where

$$\delta X(t) = \begin{bmatrix} \delta p^n \\ \delta v^n \\ \psi^n \\ \delta m_1^n \\ \vdots \\ \delta m_k^n \end{bmatrix} \quad F = \begin{bmatrix} 0 & I_{3\times 3} & 0 & 0 & \cdots & 0 \\ 0 & 0 & [\hat{f}^n\times] & 0 & \cdots & 0 \\ 0 & 0 & 0 & 0 & \cdots & 0 \\ 0 & 0 & 0 & 0 & \cdots & 0 \\ \vdots & \vdots & \vdots & \vdots & \cdots & \vdots \\ 0 & 0 & 0 & 0 & \cdots & 0 \end{bmatrix}$$

$w(t)$ is modeled as Gaussian white noise processes and $[\hat{f}^n\times]$ is the skew-symmetric matrix of the specific force vector in the n-frame.

### 2.2 SLAM observation model

$$\delta P_{mib}^n = H\delta X + v(t) \quad (2)$$

where

$$H = \begin{bmatrix} -I_{3\times 3} & 0 & [\hat{r}_{m1v}^n\times] & \cdots & I_{3\times 3} & \cdots \end{bmatrix}$$

$\delta P_{mib}^n$ is the difference between the observation and the estimated value of the ith feature, $v(t)$ is modeled as Gaussian white noise measurements and $[\hat{r}_{m1v}^n\times]$ is the skew-symmetric matrix of the estimate of the relative position between the feature and vehicle positions in the navigation frame.

## 3. Continuous PWCS Observability

The continuous PWCS observability method is introduced in [6], and we briefly show it again here for clarity.

Considering the PWCS model

$$\begin{aligned} \dot{X}(t) &= F_j X(t) \\ Z_j(t) &= H_j X(t) \end{aligned} \quad (3)$$

where

$$X(t) \in R^n; F_j \in R^{n\times n}; Z_j(t) \in R^m; H_j \in R^{m\times n}$$

The local observability matrix(LOM) is[1]

$$Q_j = \begin{bmatrix} H_j^T & (H_jF_j)^T & (H_jF_j^2)^T & \cdots & (H_jF_j^{n-1})^T \end{bmatrix}^T \quad (4)$$

If $Q_j$ has rank n, then the system defined above is locally observable.

Let $\Delta_j, j=1,2,\cdots,r$ be the time interval between $t_j$ to $t_{j+1}$ then obviously

$$X(t_{j+1}) = e^{F_j\Delta_j} X(t_j) \quad (5)$$

and the total observability matrix(TOM) is [1]

$$Q(r) = \begin{bmatrix} Q_1 \\ Q_2 e^{F_1\Delta_1} \\ \vdots \\ Q_r e^{F_{r-1}\Delta_{r-1}}\cdots e^{F_1\Delta_1} \end{bmatrix} \quad (6)$$

If $Q(r)$ has rank n, then the system defined above is global observable.

## 4. SLAM Observability

In this section, we will investigate the observability of SLAM system. Noting that in the continuous PWCS model, the dimensions of process and measurement models are fixed and the variables are the same all the time, however, in the SLAM system, we may detect different features during different segments, moreover, the former features may be not detected in the next segment, which is the fact in most SLAM system. So the dimension and the variables of the system will change during the segments. If we just concern the local observability of SLAM, well, PWCS method can be used directly, while if we are interested in the total observability of SLAM, directly using the PWCS method might be wrong and we should make some changes of the SLAM system.

Considering the next two systems

$$\Sigma_1 : \dot{X}(t) = A(t)X(t) \\ y(t) = C(t)X(t) \quad (7)$$

$$\Sigma_2 : \begin{bmatrix} \dot{X}(t) \\ 0 \end{bmatrix} = \begin{bmatrix} A(t) & 0 \\ 0 & 0 \end{bmatrix} \begin{bmatrix} X(t) \\ b \end{bmatrix} \\ \begin{bmatrix} y(t) \\ 0 \end{bmatrix} = \begin{bmatrix} C(t) & 0 \\ 0 & 0 \end{bmatrix} \begin{bmatrix} X(t) \\ b \end{bmatrix} \quad (8)$$

where

$X(t) \in R^n; A(t) \in R^{n \times n}; y(t) \in R^m; C(t) \in R^{m \times n}$,

and $b \in R^{k \times 1}$ is random vector.

It is obviously that the two systems are equivalent, that is

$$\Sigma_1 \Leftrightarrow \Sigma_2$$

So the observability of the two systems are equivalent, that is

Local observability of $\Sigma_1$ $\Leftrightarrow$ Local observability of $\Sigma_2$

Total observability of $\Sigma_1$ $\Leftrightarrow$ Total observability of $\Sigma_2$

Now we will analysis the observability of the SLAM system during the time interval $[t_0, t_k]$. Assuming that we detect $N_j$ features inside the $j$th segment, and times of the repeated features is M, the time-line of the measurements is shown in Fig.1.

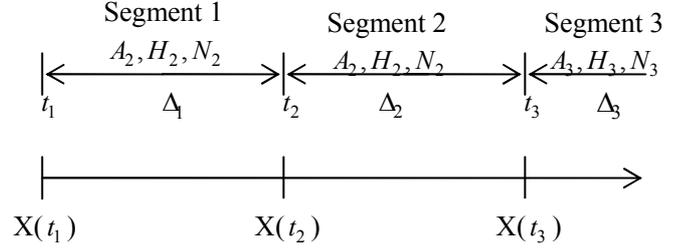

Fig.1 Measurement time-line of PWCS of SLAM

Noting that any detected feature does not move, that is $\delta \dot{m}_i^n \equiv 0$, and the number of the total detected features is $\sum_{j=1}^{k} N_j - M$, then the system model of the $i$th segment can be augmented as follows

$$\begin{bmatrix} \dot{X}(t) \\ \delta \dot{m}_1(t) \\ \vdots \\ \delta \dot{m}_{\sum_{j=1}^{k} N_j - M}(t) \end{bmatrix} = A_i \begin{bmatrix} X(t) \\ \delta m_1(t) \\ \vdots \\ \delta m_{\sum_{j=1}^{k} N_j - M}(t) \end{bmatrix} \quad (9)$$

$$= \begin{bmatrix} F_i & 0 & \cdots & 0 \\ 0 & 0 & \cdots & 0 \\ \vdots & \vdots & \ddots & \vdots \\ 0 & 0 & 0 & 0 \end{bmatrix} \begin{bmatrix} X(t) \\ \delta m_1(t) \\ \vdots \\ \delta m_{\sum_{j=1}^{k} N_j - M}(t) \end{bmatrix}$$

$$Z(t) = H_i \begin{bmatrix} X(t) \\ \delta m_1(t) \\ \vdots \\ \delta m_{\sum N_j - M}(t) \end{bmatrix}$$

$$= \begin{bmatrix} a_{1i}C_{1i}(t) & a_{1i}I_{3 \times 3} & 0 & \cdots & 0 \\ a_{2i}C_{2i}(t) & 0 & a_{2i}I_{3 \times 3} & \cdots & 0 \\ \vdots & \vdots & \vdots & \ddots & \vdots \\ a_{(\sum N_j - M)i}C_{(\sum N_j - M)i}(t) & 0 & \cdots & \cdots & a_{(\sum N_j - M)i}I_{3 \times 3} \end{bmatrix} \begin{bmatrix} X(t) \\ \delta m_1(t) \\ \vdots \\ \delta m_{\sum N_j - M}(t) \end{bmatrix}$$

(10)

where

$$X = \begin{bmatrix} \delta p^n \\ \delta v^n \\ \psi^n \end{bmatrix}, \quad F_i = \begin{bmatrix} 0 & I_{3\times 3} & 0 \\ 0 & 0 & [\hat{f}_i^n \times] \\ 0 & 0 & 0 \end{bmatrix},$$

$$C_{ci} = \begin{bmatrix} -I_{3\times 3} & 0 & [\hat{r}_{ci}^n \times] \end{bmatrix} (c = 1, \cdots, \sum_{j=1}^{k} N_j - M)$$

and $a_{ci} = 1$ if the $c$th feature is detected in the $i$th segment, otherwise $a_{ci} = 0$.

Now, we have reorganized the system model of SLAM, and the system's dimension and the variables keep steady during the total time $[t_0, t_k]$, then we can use the PWCS observability method to analyze the SLAM system.

Noting that $X(t_{i+1}) = e^{F_i \Delta_i} X(t_i)$, then

$$\begin{bmatrix} X(t_{i+1}) \\ \delta m_1(t_{i+1}) \\ \vdots \\ \delta m_{\sum_{j=1}^{k} N_j - M}(t_{i+1}) \end{bmatrix} = \begin{bmatrix} e^{F\Delta} & 0 & \cdots & 0 \\ 0 & I_{3\times 3} & \cdots & 0 \\ \vdots & \vdots & \ddots & \vdots \\ 0 & 0 & \cdots & I_{3\times 3} \end{bmatrix} \begin{bmatrix} X(t_i) \\ \delta m_1(t_i) \\ \vdots \\ \delta m_{\sum_{j=1}^{k} N_j - M}(t_i) \end{bmatrix}$$

(11)

Similar with [1], the TOM of SLAM system is

$$Q(k) = \begin{bmatrix} H_1 \\ H_1 A_1 \\ \vdots \\ H_1 A_1^{N_1+8} \\ \hline H_2 \begin{bmatrix} e^{F_1 \Delta_1} & 0 & \cdots & 0 \\ 0 & I_{3\times 3} & \cdots & 0 \\ \vdots & \vdots & \ddots & \vdots \\ 0 & 0 & \cdots & I_{3\times 3} \end{bmatrix} \\ H_2 A_2 \begin{bmatrix} e^{F_1 \Delta_1} & 0 & \cdots & 0 \\ 0 & I_{3\times 3} & \cdots & 0 \\ \vdots & \vdots & \ddots & \vdots \\ 0 & 0 & \cdots & I_{3\times 3} \end{bmatrix} \\ \vdots \\ H_2 A_2^{N_2+8} \begin{bmatrix} e^{F_1 \Delta_1} & 0 & \cdots & 0 \\ 0 & I_{3\times 3} & \cdots & 0 \\ \vdots & \vdots & \ddots & \vdots \\ 0 & 0 & \cdots & I_{3\times 3} \end{bmatrix} \\ \hline \vdots \\ \hline H_k \begin{bmatrix} e^{F_{k-1}\Delta_{k-1}} e^{F_{k-2}\Delta_{k-2}} \cdots e^{F_1 \Delta_1} & 0 & \cdots & 0 \\ 0 & I_{3\times 3} & \cdots & 0 \\ \vdots & \vdots & \ddots & \vdots \\ 0 & 0 & \cdots & I_{3\times 3} \end{bmatrix} \\ H_k A_k \begin{bmatrix} e^{F_{k-1}\Delta_{k-1}} e^{F_{k-2}\Delta_{k-2}} \cdots e^{F_1 \Delta_1} & 0 & \cdots & 0 \\ 0 & I_{3\times 3} & \cdots & 0 \\ \vdots & \vdots & \ddots & \vdots \\ 0 & 0 & \cdots & I_{3\times 3} \end{bmatrix} \\ \vdots \\ H_k A_k^{N_k+8} \begin{bmatrix} e^{F_{k-1}\Delta_{k-1}} e^{F_{k-2}\Delta_{k-2}} \cdots e^{F_1 \Delta_1} & 0 & \cdots & 0 \\ 0 & I_{3\times 3} & \cdots & 0 \\ \vdots & \vdots & \ddots & \vdots \\ 0 & 0 & \cdots & I_{3\times 3} \end{bmatrix} \end{bmatrix}$$

(12)

## 5. Observability Rank Analysis

Before the observability rank analysis, we assume there are L features detected during the time interval $[t_0, t_k]$, and the features are signed as $\{1, 2, \cdots, L\}$.

In all of the following observability matricies, we only present the non-zero rows, and have removed those rows that contain only zeros as they do not contribute to the observability analysis. Let the time interval between $t_j$ to $t_{j+1}$ $\Delta_j = \Delta t, j = 1, 2, \cdots, r$, and the matrix is expanded using only the first-order term, i.e., $e^{F\Delta t} = I + F\Delta t$.

Substituting Eqs.(9-10) to Eq.(12) yields

$$Q(k) = \begin{bmatrix} -a_{11}I_{3\times 3} & 0 & a_{11}[\hat{r}_{11}^n \times] & a_{11}I_{3\times 3} & \cdots & 0 \\ \vdots & \vdots & \vdots & \vdots & \ddots & \vdots \\ -a_{L1}I_{3\times 3} & 0 & a_{L1}[\hat{r}_{L1}^n \times] & 0 & \cdots & a_{L1}I_{3\times 3} \\ 0 & -a_{11}I_{3\times 3} & 0 & 0 & \cdots & 0 \\ \vdots & \vdots & \vdots & \vdots & \ddots & \vdots \\ 0 & -a_{L1}I_{3\times 3} & 0 & 0 & \cdots & 0 \\ 0 & 0 & -a_{11}[\hat{f}_1^n \times] & 0 & \cdots & 0 \\ \vdots & \vdots & \vdots & \vdots & \ddots & \vdots \\ 0 & 0 & -a_{L1}[\hat{f}_1^n \times] & 0 & \cdots & 0 \\ \hline -a_{12}I_{3\times 3} & -a_{12}I_{3\times 3}\Delta t & a_{12}[\hat{r}_{12}^n \times] & a_{12}I_{3\times 3} & \cdots & 0 \\ \vdots & \vdots & \vdots & \vdots & \ddots & \vdots \\ -a_{L2}I_{3\times 3} & -a_{L2}I_{3\times 3}\Delta t & a_{L2}[\hat{r}_{L2}^n \times] & 0 & \cdots & a_{L2}I_{3\times 3} \\ 0 & -a_{12}I_{3\times 3} & -a_{12}\Delta t[\hat{f}_1^n \times] & 0 & \cdots & 0 \\ \vdots & \vdots & \vdots & \vdots & \ddots & \vdots \\ 0 & -a_{L2}I_{3\times 3} & -a_{L2}\Delta t[\hat{f}_1^n \times] & 0 & \cdots & 0 \\ 0 & 0 & -a_{12}[\hat{f}_2^n \times] & 0 & \cdots & 0 \\ \vdots & \vdots & \vdots & \vdots & \ddots & \vdots \\ 0 & 0 & -a_{L2}[\hat{f}_2^n \times] & 0 & \cdots & 0 \\ \hline -a_{13}I_{3\times 3} & -2a_{13}I_{3\times 3}\Delta t & a_{13}[\hat{r}_{13}^n \times] & a_{13}I_{3\times 3} & \cdots & 0 \\ \vdots & \vdots & \vdots & \vdots & \ddots & \vdots \\ -a_{L3}I_{3\times 3} & -2a_{L3}I_{3\times 3}\Delta t & a_{L3}[\hat{r}_{L3}^n \times] & 0 & \cdots & a_{L3}I_{3\times 3} \\ 0 & -a_{13}I_{3\times 3} & -a_{13}\Delta t([\hat{f}_1^n \times]+[\hat{f}_2^n \times]) & 0 & \cdots & 0 \\ \vdots & \vdots & \vdots & \vdots & \ddots & \vdots \\ 0 & -a_{L3}I_{3\times 3} & -a_{L3}\Delta t([\hat{f}_1^n \times]+[\hat{f}_2^n \times]) & 0 & \cdots & 0 \\ 0 & 0 & -a_{13}[\hat{f}_3^n \times] & 0 & \cdots & 0 \\ \vdots & \vdots & \vdots & \vdots & \ddots & \vdots \\ 0 & 0 & -a_{L3}[\hat{f}_3^n \times] & 0 & \cdots & 0 \\ \hline \vdots \\ \hline -a_{1k}I_{3\times 3} & -(k-1)a_{1k}I_{3\times 3}\Delta t & a_{1k}[\hat{r}_{1k}^n \times] & a_{1k}I_{3\times 3} & \cdots & 0 \\ \vdots & \vdots & \vdots & \vdots & \ddots & \vdots \\ -a_{Lk}I_{3\times 3} & -(k-1)a_{Lk}I_{3\times 3}\Delta t & a_{Lk}[\hat{r}_{Lk}^n \times] & 0 & \cdots & a_{Lk}I_{3\times 3} \\ 0 & -a_{1k}I_{3\times 3} & -a_{1k}\Delta t(\sum_{i=1}^{k-1}[\hat{f}_i^n \times]) & 0 & \cdots & 0 \\ \vdots & \vdots & \vdots & \vdots & \ddots & \vdots \\ 0 & -a_{Lk}I_{3\times 3} & -a_{Lk}\Delta t(\sum_{i=1}^{k-1}[\hat{f}_i^n \times]) & 0 & \cdots & 0 \\ 0 & 0 & -a_{1k}[\hat{f}_k^n \times] & 0 & \cdots & 0 \\ \vdots & \vdots & \vdots & \vdots & \ddots & \vdots \\ 0 & 0 & -a_{Lk}[\hat{f}_k^n \times] & 0 & \cdots & 0 \end{bmatrix}$$

(13)

Without loss of generality, taking two segments for example, then the TOM is

$$Q(2)=\begin{bmatrix} -a_{l1}I_{3\times3} & 0 & a_{l1}[\hat{r}_{11}^n\times] & a_{l1}I_{3\times3} & \cdots & 0 \\ \vdots & \vdots & \vdots & \vdots & \ddots & \vdots \\ -a_{L1}I_{3\times3} & 0 & a_{L1}[\hat{r}_{L1}^n\times] & 0 & \cdots & a_{L1}I_{3\times3} \\ 0 & -a_{l1}I_{3\times3} & 0 & 0 & \cdots & 0 \\ \vdots & \vdots & \vdots & \vdots & \ddots & \vdots \\ 0 & -a_{L1}I_{3\times3} & 0 & 0 & \cdots & 0 \\ 0 & 0 & -a_{l1}[\hat{f}_1^n\times] & 0 & \cdots & 0 \\ \vdots & \vdots & \vdots & \vdots & \ddots & \vdots \\ 0 & 0 & -a_{L1}[\hat{f}_1^n\times] & 0 & \cdots & 0 \\ -a_{l2}I_{3\times3} & -a_{l2}I_{3\times3}\Delta t & a_{l2}[\hat{r}_{l2}^n\times] & a_{l2}I_{3\times3} & \cdots & 0 \\ \vdots & \vdots & \vdots & \vdots & \ddots & \vdots \\ -a_{L2}I_{3\times3} & -a_{L2}I_{3\times3}\Delta t & a_{L2}[\hat{r}_{L2}^n\times] & 0 & \cdots & a_{L2}I_{3\times3} \\ 0 & -a_{l2}I_{3\times3} & -a_{l2}\Delta t[\hat{f}_1^n\times] & 0 & \cdots & 0 \\ \vdots & \vdots & \vdots & \vdots & \ddots & \vdots \\ 0 & -a_{L2}I_{3\times3} & -a_{L2}\Delta t[\hat{f}_1^n\times] & 0 & \cdots & 0 \\ 0 & 0 & -a_{l2}[\hat{f}_2^n\times] & 0 & \cdots & 0 \\ \vdots & \vdots & \vdots & \vdots & \ddots & \vdots \\ 0 & 0 & -a_{L2}[\hat{f}_2^n\times] & 0 & \cdots & 0 \end{bmatrix}$$

(14)

We can further simplify the observation problem by assuming the detected features L=2. Then there might be four cases listed in Table 1.

Table 1 Four cases during two segments

| case | segment | |
|---|---|---|
| | 1 | 2 |
| 1 | feature 1 | feature 2 |
| 2 | feature 1 | feature 1,2 |
| 3 | feature 1,2 | feature 2 |
| 4 | feature 1,2 | feature 1,2 |

Taking case 2 for example, the LOM of case 2 during segment 1 is

$$Q_{LOM} = \begin{bmatrix} -I_{3\times3} & 0 & [\hat{r}_{m1}^n\times] & I_{3\times3} \\ 0 & -I_{3\times3} & 0 & 0 \\ 0 & 0 & -[\hat{f}^n\times] & 0 \end{bmatrix} \quad (15)$$

Given that $\hat{f}^n \neq 0$, the LOM in (15) has a rank of eight where the total number of estimated states is twelve. Only the velocity $\delta v^n$ is observable, and the others states, such as position $\delta p^n, \delta m^n$, attitude $\psi^n$ are not observable.

After two segments, the TOM of the system is

$$Q_{TOM} = \begin{bmatrix} -I_{3\times3} & 0 & [\hat{r}_{11}^n\times] & I_{3\times3} & 0 \\ -I_{3\times3} & 0 & [\hat{r}_{21}^n\times] & 0 & I_{3\times3} \\ 0 & -I_{3\times3} & 0 & 0 & 0 \\ 0 & -I_{3\times3} & 0 & 0 & 0 \\ 0 & 0 & -[\hat{f}_1^n\times] & 0 & 0 \\ 0 & 0 & -[\hat{f}_1^n\times] & 0 & 0 \\ -I_{3\times3} & -I_{3\times3}\Delta t & [\hat{r}_{22}^n\times] & 0 & I_{3\times3} \\ 0 & -I_{3\times3} & -\Delta t[\hat{f}_1^n\times] & 0 & 0 \\ 0 & 0 & -[\hat{f}_2^n\times] & 0 & 0 \end{bmatrix}$$

(16)

Provided that $\hat{f}_1^n \times \hat{f}_2^n \neq 0$, the TOM in (16) has a rank of twelve, where the total number of estimated states is fifteen, so the system is not observable during the time interval $[t_0, t_2]$, and the observable modes are $\delta v^n, \psi^n, \delta p^n - \delta m_2^n, \delta m_1^n - \delta m_2^n$.

Similarly, we can analyze the observability of other three cases, and the results of case 1 and 4 are the same as [6].

In general, we can analyze the observability of any case using (13), if considering one segment, then the TOM in (13) is reduced to the LOM case. However regardless of the number of features added to the map or any extra time segments, the number of unobservable states will never be less than three.

## 6. COVARIANCE SIMULATION RESULTS

In this section, some simulations will be done on a computer with Pentium® Dual-Core CPU and 2GB memory. The code is programmed using MATLAB 2009.

A low-cost IMU is simulated with a vision as the range, bearing and elevation sensor. The simulation parameters are listed in Table 2.

Table 2. The parameters used in simulation

| Sensor | specification | parameter |
|---|---|---|
| IMU | Sampling rate($Hz$) | 100 |
| | Accel noise($m/s^2$) | 0.01 |
| | Gyro noise($^o/s$) | 0.1 |
| Vision | Frame rate($Hz$) | 25 |
| | Field-of-view($^o$) | ±15 |
| | range error(m) | 5 |
| | Bearing noise($^o$) | 0.1 |
| | Elevation noise($^o$) | 0.1 |

The major motivation for the observability analysis is the correspondence between this analysis and the estimability of the system. The estimability of the UAV system was checked using a covariance simulation, as it predicts the performance of a Kalman filter designed for the system.

Assuming all the errors in the process and sensor to be Gaussian white, numerical simulation results on the standard deviation (STD) of errors are provided using extended kalman filter(EKF) in the following. In the simulation, the covariance simulation was carried out in a piece-wise constant acceleration trajectory listed in Table 3.

Table 3 Segments of Accelerations

| segment | accelerations | | |
|---|---|---|---|
| | $F_N$ | $F_E$ | $F_U$ |
| 1 | 0 | 0 | g |
| 2 | 0 | 0.1 | g |

The trajectory consisted of segements 1,2 of Table 3 and in this order. Each segment was run for 50s.

The vehicle motion with the origin location at $P_0 = [0,0,100]^T$, The initial velocity is given by $V_0 = [0.1,0,0]^T$. The initial covariance matrix P for the vehicle is set as follows

$P = \text{diag}([1,1,1,1,1,1,0.0873,0.0873,0.0873])$

Similarly with [11], the copy of the covariance matrix $U_m$ is augmented to include the initial covariance of the features for convenience

$$P_{aug} = \begin{bmatrix} P_{xx} & P_{xm} & 0 \\ P_{mx} & P_{mm} & 0 \\ 0 & 0 & U_m \end{bmatrix} \quad (17)$$

The covariance of the expected feature positions before any observations $U_m$ is given an large initial value( diagonal matrix with values of $10^9 m^2$ along the diagonal)

In the simulation, the true map features are $F_1 = [10,0,0]^T, F_2 = [20,100,0]^T$.

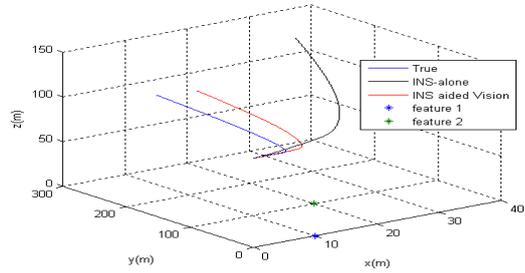

Fig.2 True, INS and estimated vehicle trajectories

From Fig.2, it can be seen that the estimated vehicle trajectory is more close to the true trajectory than the INS alone trajectory.

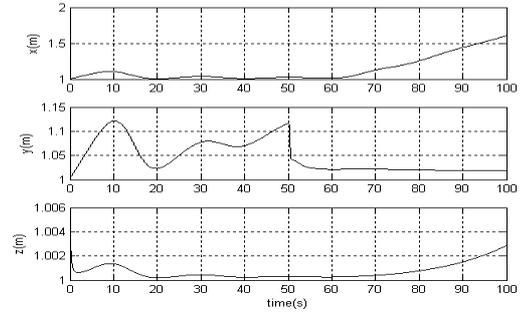

Fig.3 Covariance simulation of position error

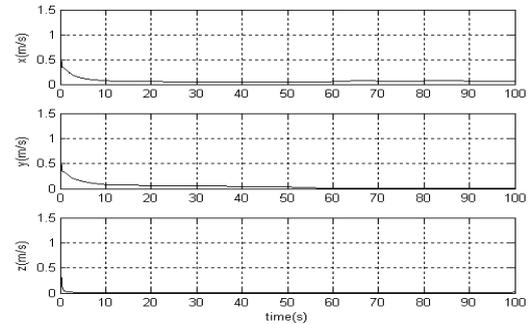

Fig.4 Covariance simulation of velocity error

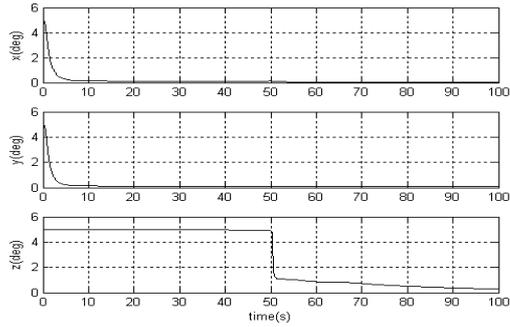

Fig.5 Covariance simulation of attitude error

Figs. 3-5 show errors in the estimates of INS states. The STDs of position error were misleading in Fig.3, while the STDs of velocity error decreased significantly at the beginning in Fig.4, this is because during the whole segments, the vehicle position is not observable, while its velocity is observable. The STDs of roll and pitch estimation errors in Fig.5 decreased significantly at the beginning. This is because the horizontal components of unobservable attitude error can be approximated to the accelerometer bias divided by gravity. The yaw estimation error during the first 50s given in Fig.6 can be misleading, and decreased during the second 50s, this is because the attitude is observable during segment 2.

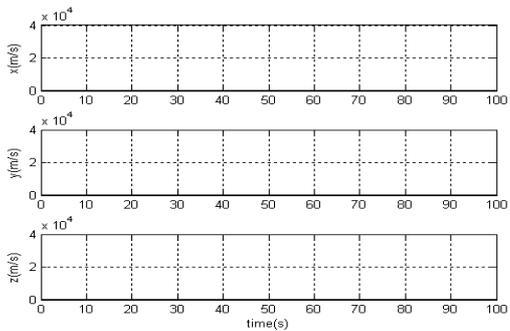

Fig.6a  Covariance simulation of feature 1

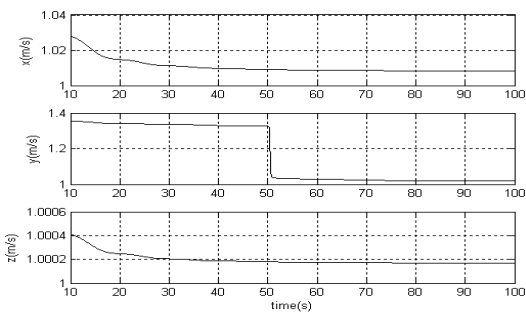

Fig.6b Covariance simulation of feature 1 in 10~100s

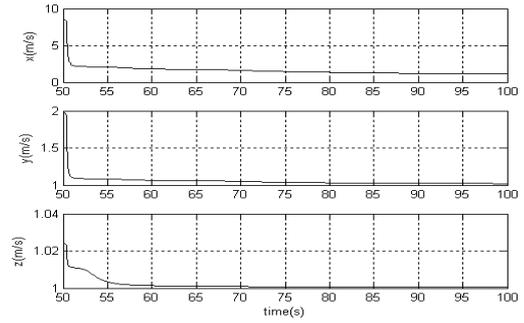

Fig.7 Covariance simulation of feature 2 in 50~100s

In Figs. 6 and 7, the two features estimated errors do not converge to zero, this is because the two features are not observable in the system.

Figs.8 and 9 show that the relative position of the feature to the vehicle and the relative position between the features estimated errors converge to zero, this is because they are observable in the system.

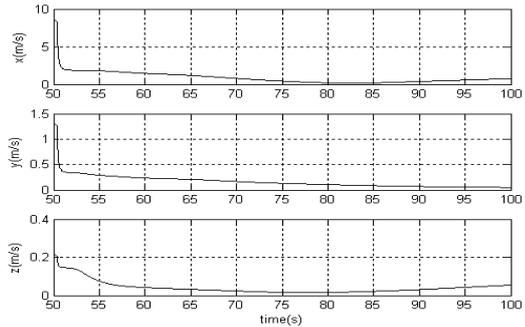

Fig.8 Covariance simulation of relative position of feature 2 to the vehicle in 50~100s

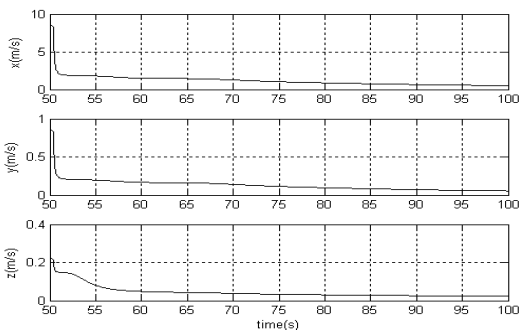

Fig.9 Covariance simulation of relative position between feature 1 and 2 in 50~100s

## VII. CONCLUSIONS

A unified approach for analyzing the observability of SLAM system is provided. Simulation results demonstrates the valid of the approach. The main contribution is that we can easily find out the observability of any case in the SLAM system using this approach, and the previous work appeared in [11] is the especial example in this paper. This conclusion might be very useful for the SLAM system design, such as the determination of the vehicle action selection.